\documentclass[conference]{IEEEtran}
\IEEEoverridecommandlockouts
\usepackage{graphicx}
\usepackage{float}    
\usepackage{url} 
\usepackage{amsmath,amssymb,amsfonts}
\usepackage{algorithmic}
\usepackage{hyperref} 
\usepackage{textcomp}
\usepackage{cite}
\usepackage{xcolor}
\usepackage{graphicx}

\def\BibTeX{{\rm B\kern-.05em{\sc i\kern-.025em b}\kern-.08em
    T\kern-.1667em\lower.7ex\hbox{E}\kern-.125emX}}
\begin{document}

\title{ An Attention Infused Deep Learning System with  Grad-CAM Visualization for Early Screening of Glaucoma\\
}

\author{\IEEEauthorblockN{Ramanathan Swaminathan}
\IEEEauthorblockA{\textit{College of Computing} \\
\textit{Georgia Institute of Technology}\\
{Atlanta, Georgia,}\\
rswaminathan38@gatech.edu}

}

\maketitle

\begin{abstract}
This research work reveals the strengths of intertwining a deep custom convolutional neural network with a disruptive Vision Transformer, both fused together with a radical Cross-Attention module. Here, two high-yielding datasets for artificial intelligence models in detecting glaucoma, namely ACRIMA and Drishti, are utilized. The Cross-Attention mechanism facilitates the model in learning regions in the fundus that are clinically relevant through bidirectional feature exchange between CNN and ViT streams. Experiments clearly depict improved performance when compared to standalone baseline CNN and ViT models. 
\end{abstract}


\section{Introduction}
The field of medical imaging has been extremely influenced and revolutionized by the advent of traditional layer-based gold-standard architectures like convolutional neural networks. But these do come with their limitations when applied to multifaceted domains like medical imaging, where infinitesimal details of the fundus images determine the difference between true positives and false positives.  

With respect to the detailed research conducted in this study, an inventive amalgamation of both the undisputed top performer in vision-based tasks, the Vision Transformers, and the old but gold convolutional neural networks is carried out. This was only achievable due to the creation of a powerful cross-attention network mechanism to combine the two highly contrasting contenders. As anticipated, according to the detailed ablation studies, the combined performance of this hybrid concoction model outperformed all the other non-hybrid and non-attention-based models.

The empowerment of this strong hybrid model with cross-attention has facilitated the process of feature weighing and calibration  to become dynamically prioritized to foster the environment for improved feature recognition, and thus a state-of-the-art model with superior generalization and deployability is born. 

The extensive and unifying motive of this research work is to support doctors in various eye care systems around the globe, especially in undeveloped and developing countries, to at least semi-automate the early diagnosis and prompt treatment of the dominant cause of preventable blindness without complications or severe loss of vision.

\section{Related Work and Literature Review}

A novel synthetic data generator called generative adversarial networks (GAN) was introduced by [1] Chaurasia et al. (2024), where synthetic optic disc and cup images were synthesized. This research lucidly indicates the hurdles in obtaining a quality fundus dataset and emphasizes self-reliance by using machine-generated, real-like images, which were even contributing to the models' improved generalization. This type of data sourcing is highly cost optimized but relies heavily on data augmentation techniques. Glaucoma was categorized into three classes—mild, moderate, and severe—by [2] Jisy NK et al. (2024), where early glaucoma detection was stressed along with feature visualization techniques to enable better model understandability. A quadruple dataset consisting of images from DRISHTI, Origa, G1020, and RIM-ONE DL was used here, but still accuracy levels of not more than 93.75\% were only achievable at the cost of improved generalizability and reduced false negatives.

[5]. Ramanathan et al. (2023) - In my previous work,  I had explored convolutional neural nets in depth. I made use of  multi-scale and region-guided attention mechanisms for detecting diabetic retinal diseases, namely  diabetic retinopathy and diabetic macular edema. These diseases  are way harder to treat once vision loss starts, unlike glaucoma. A comprehensive multimodal dataset from Instituto Mexicano de Oftalmología was leveraged. [6]. Ajitha et al. (2021) produced a robust CNN model consisting of 13 layers trained and tested over a massive 1,113-image dataset with extensive efforts for data augmentation as well. It is highly notable that they used both SVM and Softmax as the final classification protocol in their CNN model. Despite such complexity, they were only able to reach up to 95.61\% accuracy. Although it is highly respectable that their precision and specificity were both sweeping 100 \%. [7]. Saha et al. (2023) worked specifically on localizing the optic nerve head,  using a lean variant of the critically acclaimed YOLO nano architecture. A MobileNetV3Small CNN model was curated, and the optic nerve head was fed to this model. The top seven public datasets were combined together to form 6,671 images. Despite using loads of images, they were able to achieve high memory efficiency levels of almost 12 times less than a traditional ResNet50 model. This will enable them to roll out their model over resource-constrained environments.

[8]. Coan et al. (2023), despite not creating a novel model, reviewed and surveyed the ways to harness the strengths of both machine learning and non-AI-rule-based models. It is outstanding that they emphasized the models that isolate and study the optic cup and disc's frontiers, as they play a prime function in evaluating glaucoma. Their paper was a concise summary of the various literatures present in current-day DL systems. [9]. Akter et al. (2022) made eminent progress in advancing the frontiers of glaucoma detection by creating a real-time dataset utilizing the optical coherence tomography (OCT) images. They had collected around 255 images, with 200 images for  training followed by 55 unseen and labeled images for validating the performance of the model. The dataset is proprietary and is not publicly available. In spite of a small dataset, which is generally prone to overfitting, they were able to reach 96\% accuracy levels. This is due to the derivation of optic nerve head (ONH) and cup features from the OCT images, and this clearly exemplifies the gravity of ONH in glaucoma-diagnosing trainable AI models.

[10]. Camara et al. (2022) conducted an exhaustive and complete review of the utilization of various artificial intelligence-based methodologies, especially for segmenting and classifying the fundus images. The optic disc and cup were treated as the most paramount of all in their study, as they are proven to be the best markers for glaucoma presence. They stressed the role of AI in facilitating cost-conscious methods and precise detection of glaucoma. 

[11]. Liu Li et al. (2019) notably produced an inventive and sophisticated convolutional neural network along with attention mechanism incorporation. It is an eye-opener from their research that most fundus images contain regions that are very identical, and it is hard to spot the difference unless they are juxtaposed. To handle this nuanced intricacy, the technical prowess of the attention mechanisms was leveraged, with which only the crucial regions of the fundus photographs were given the overriding importance. 

[12]. Ashtari Majlan et al. (2023) have given enlightening and thought-provoking insights on various methodologies that are leveraged as of now. Mainly,  classification models, hybrid and ensemble models, and segmentation-based models were explicitly discussed in depth. Time and again the optic disc and cup segmentation's criticality is reiterated along with dataset annotation for model performance like never before. The black-box kind of AI models remain a huge impediment in terms of explainable AI, and they are stressing about improving the transparency of the research work to make sure more of the general audience and investors gain trust with these complex modern tools. The plethora of drawbacks, like fundus photo quality and lighting issues, along with class imbalance handling, is exhaustively covered in their research study.

 A seamless and integrated Transfer Induced Attention Network (TIANet) model was proposed by [13] Xu et al. (2021). This TIANet model is a concoction of both transfer learning via the exemplary ResNet50 model and attention mechanism with special detail to the optic cup and disc clusters. It is important to make note of the fact that the RIM-ONE-r3 was made use of here and the TIANet model performed fairly well with 96.7\% accuracy. In addition to all this, attention maps were created to interpret how the model prioritized the feature selection .

An unbiased and fair system for the sole purpose of screening glaucoma disease was created by [14] Shi et al. No matter the age, whether young or old, race, or gender, there is no sort of prejudice that is invoked by the model.

An idiosyncratic and one-of-a-kind glaucoma detection system was put into use by [15] Fan et al. It is important to note that this model achieved impressive generalization despite the absence of any convolution layers. A DeiT model, which is a data-efficient image transformer, was used here that has only self-attention mechanisms incorporating parts like a multi-head attention module and a coherent feedforward network with necessary normalizations.

\section{Methodology}

\subsection{Overview}

In simple terms, glaucoma, with time, will lead to irreversible damage to the optic nerve. During the progression of this disease, the patients will be able to sense losses in their peripheral vision. The methodology deployed in this research mainly was determined with respect to the type of images that were used to program the model. The images contain loads of spatial features, and the time-tested, proven deep learning model for this is obviously convolutional neural networks. This is the intent behind the use of EfficientNet-B0. Vision Transformers are among the best for vision-oriented medical imaging tasks.

Merging EfficientNet-B0 with Vision Transformers (ViT) is a one plus one is more than two approach.

\subsection{Dataset Preparation}

A hybrid dataset was created by conjoining two different fundus datasets. [14] Drishti is an open-source dataset from IIIT-Hyderabad with origins from Aravind Eye Hospital in India. The [15] ACRIMA dataset was introduced for the ACRIMA challenge by the University of Alicante, Spain. The ACRIMA dataset is distributed in an 80:20 ratio across training and testing sections. Whereas, the Drishti dataset is evenly dispersed with a 50:50 proportion.

\begin{table}[htbp]
\caption{Breakdown of Dataset Distribution}
\centering
\begin{tabular}{|p{3cm}|c|c|c|}
\hline
\textbf{Dataset} & \textbf{Split} & \textbf{Normal} & \textbf{Glaucoma} \\
\hline
\textbf{ACRIMA} & Training & 248 & 318 \\
\cline{2-4}
& Testing & 63 & 80 \\
\hline
\textbf{Drishti} & Training & 19 & 33 \\
\cline{2-4}
& Testing & 14 & 39 \\
\hline
\end{tabular}
\end{table}

\begin{table}[htbp]
\caption{Combined dataset details}
\centering
\begin{tabular}{|p{3cm}|c|c|c|}
\hline
\textbf{Split} & \textbf{Normal} & \textbf{Glaucoma} & \textbf{Total} \\
\hline
Training & 267 & 351 & 618 \\
\hline
Testing & 77 & 119 & 196 \\
\hline
\textbf{Overall} & 344 & 470 & 814 \\
\hline
\end{tabular}
\end{table}

The Drishti dataset is comparably tiny to ACRIMA, but the addition of these images is an excellent way to generalize and use the model for countries like India, as Drishti is an Indian eye fundus image dataset.

The total number of fundus images capable of showing manifestations of glaucoma was 814. In this dataset, 344 images were normal and 470 images were glaucomatous, indicating a mild class imbalance that favors the class glaucoma. The training set had wholly 618 images, while testing was just a fraction of around 1/3 of the training images with 196 images.

To make sure the memory constraints of the training device were complied with, the images were downsized to the standard 224*224 size. Furthermore, normalization of pixel values between zero and one was performed. With generalizability as a major goal, these two datasets were combined together as they are.

\subsection{Preprocessing and Augmentation}

Complex glaucoma detection models call for the need for top-tier, noise-free datasets. No dataset that is publicly available is inherently noiseless and spotlessly clean. Therefore, strong data pre-processing was carried out, as well as image augmentation, to handle class imbalances and to synthetically increase the image input such that the data-hungry models can train well. One-hot encoding was done to represent 1 with glaucoma and 0 with non-glaucoma. Data augmentation was performed via color jitter, random horizontal flip, random affine transformations and rotations, Gaussian blur for noise addition, and random resized crop.

\subsection{Model Architecture}

A fusion hybrid model comprising both Vision Transformers (ViT) and EfficientNet-B0, which is an efficient convolutional neural network, has been made use of here. The merits of these two models complement each other very well, as both local and global feature extraction are realized using this fusion model. These efforts are clearly evident via the model's improved ability to generalize over unseen data.

This proposed model consists of the following parts:
\begin{itemize}
    \item A backbone CNN model - EfficientNet-B0
    \item Vision Transformer (ViT)
    \item Cross-Attention mechanism
    \item A fully connected layer
    \item A binary output classification layer
\end{itemize}

\vspace{0.5cm}
\begin{figure}[H]
    \centering
    \includegraphics[width=1.0\linewidth]{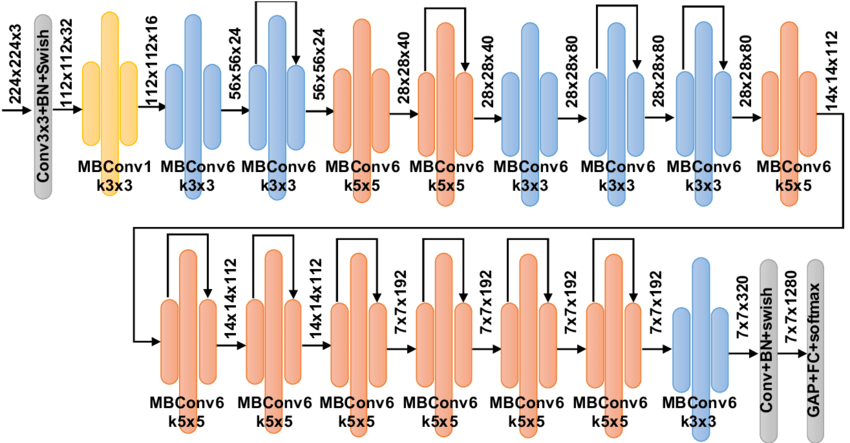}
    \caption{[16] The various layers of the EfficientNet-B0 model}
    
\end{figure}

EfficientNet-B0, as the name suggests, is tuned for efficiency via a plethora of factors like neural network depth, width, and image pixel quantity. Despite the large fundus image, the main factors that determine the presence of glaucoma are mostly the areas in and around the optic disc and cup, optic nerve thickness, along with any active bleeding. All these are very nuanced and subtle characteristics of the fundus images, and EfficientNet-B0 excels in detecting all these. EfficientNet-B0 also outperforms most models when it comes to dealing with class imbalances and limited real-time labeled data.  But it is wise to keep in mind that most CNN models lack the global abstracted context due to fairly limited spatial sensitivity. CNNs have strong, ready-made rules on prioritization of the pixels in the image, so it is to be understood that CNNs do have their own inductive biases that need to be neutralized for the model with the best caliber to be born.
\begin{figure}[H]
    \centering
    \includegraphics[width=1.0\linewidth]{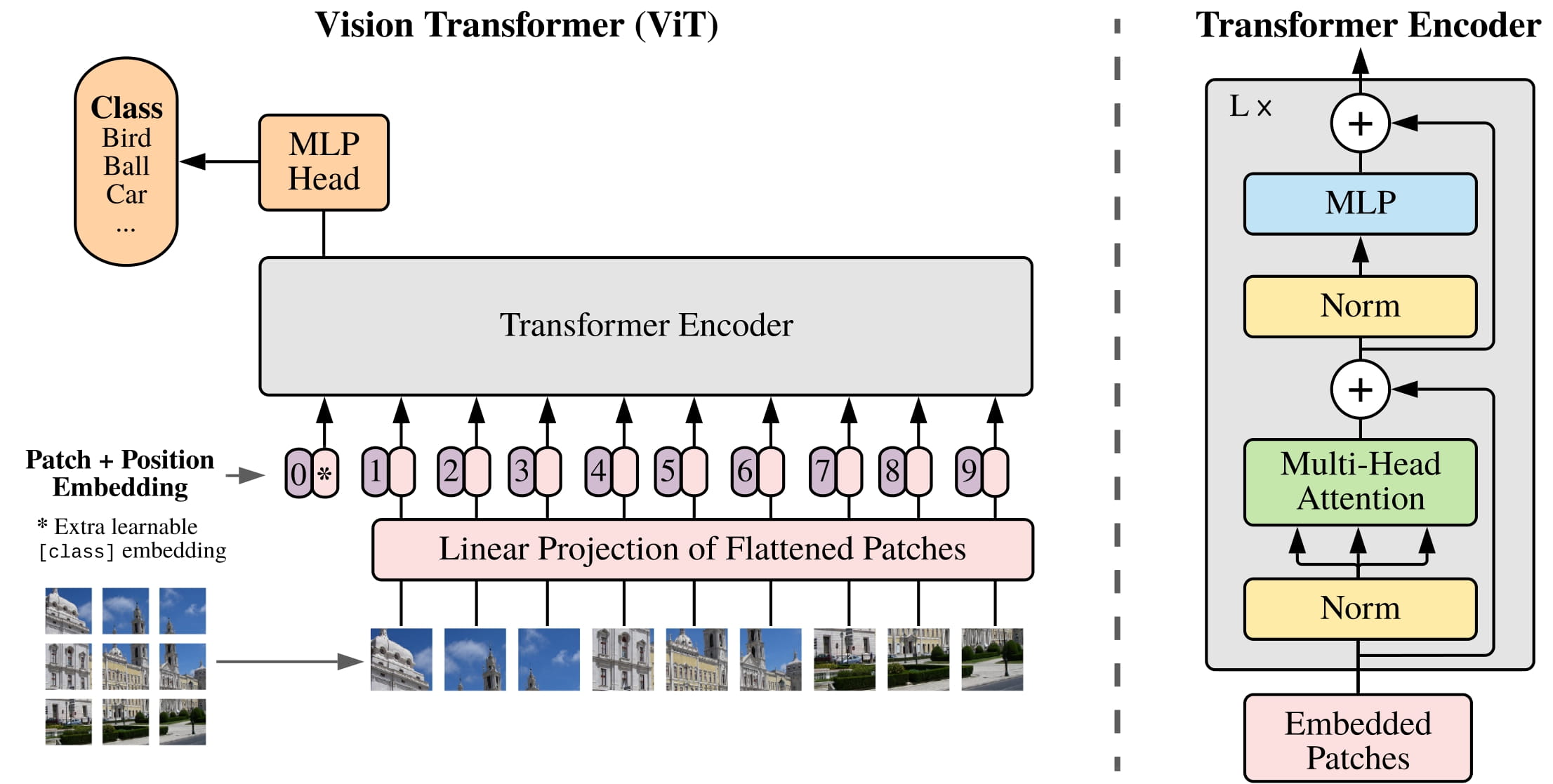}
    \caption{[17] Comprehensive architecture of Vision Transformers is clearly depicted.}
\end{figure}

On the contrary, the Vision Transformer (ViT model) models the global context exceptionally well. This is primarily due to the presence of self-attention layers that weigh the diverse, far-flung parts of the image appropriately. Vision Transformers also enable feature visualization via their attention maps. Unlike EfficientNet-B0, it does not have any significant inductive biases and can perform selective dynamic allocation of the pixels in the image. But even this sophisticated model does not come without its bottlenecks and pitfalls. Vision Transformers' thirst for data is almost unquenchable; as more data is fed, the better the model's throughput. These are not the most memory-efficient models, especially when compared with the EfficientNet-B0 model, and are vulnerable to minute fine details in the image, like minor bleeding or changes in the nerves' thickness.

The cross-attention mechanism handles both the different modalities by an adaptive feature-weighing process, which ends up combining both these modalities to create a fused feature. The cross-attention module has two major input feeds, which include
\begin{itemize}
    \item First and foremost, the output from the CNN-EfficientNet-B0 model.
    \item Secondly, the ViT model's output is also fed as the secondary input for the cross-attention mechanism.
    
\end{itemize}

A dense, fully connected layer is utilized for dimensionality reduction of the combined feature set. This is further passed on to the final binary classifier layer that has a sigmoid activation function that enables classification as either glaucomatous or normal. The binary cross entropy loss function along with the Adam optimizer is implemented here. A dynamic automated learning rate scheduler called CosineAnnealingLR is used to find the optimal learning rate.

\subsection{Training Protocol}

The entire training process was made possible due to the powerful Apple Silicon M3 Pro CPU with the 18-core GPU acceleration via Metal Performance Shaders (MPS). The learning rate was set to 5e-4 and supplemented with a weight decay of 1e-4. The number of epochs was set to 50, and the batch size was reduced to 16 in order to handle the 18 GB GPU memory limitation. Gradient clipping was also introduced with a maxnorm value of 1.0. The automated model checkpointing made sure the best model after every epoch was saved. The various gold-standard evaluation metrics, like the confusion matrix, training accuracy, and testing accuracy, along with the classification report, were generated for perceiving model's performance.

\section{Results and Discussion}
\subsection{Performance Evaluation}

The detailed performance statistics of the proposed hybrid model have been lucidly stated in Table III, which clearly depicts how effortlessly it is able to find the contrasts between healthy and glaucoma-affected eyes. Its performance is highlighted by the 94.8\% accuracy score. The various other parameters like F1-score, precision, and recall are also top-notch. The weighted average and macro averages are also highly respectable, meaning that the model performs well over both classes and has almost no trouble handling class imbalances.

\begin{table}[ht]
\centering
\caption{Classification Report }
\begin{tabular}{|l|c|c|c|c|}
\hline
\textbf{Class} & \textbf{Precision} & \textbf{Recall} & \textbf{F1-score} & \textbf{Support} \\
\hline
Glaucoma       & 0.950 & 0.966 & 0.958 & 117 \\
Normal         & 0.945 & 0.920 & 0.933 & 75  \\
\hline
\textbf{Accuracy}     &       &       & \textbf{0.948} & 192 \\
\textbf{Macro avg}    & 0.947 & 0.943 & 0.946 & 192 \\
\textbf{Weighted avg} & 0.948 & 0.948 & 0.948 & 192 \\
\hline
\end{tabular}
\end{table}

\begin{figure}[H]
    \centering
    \includegraphics[width=1.0\linewidth]{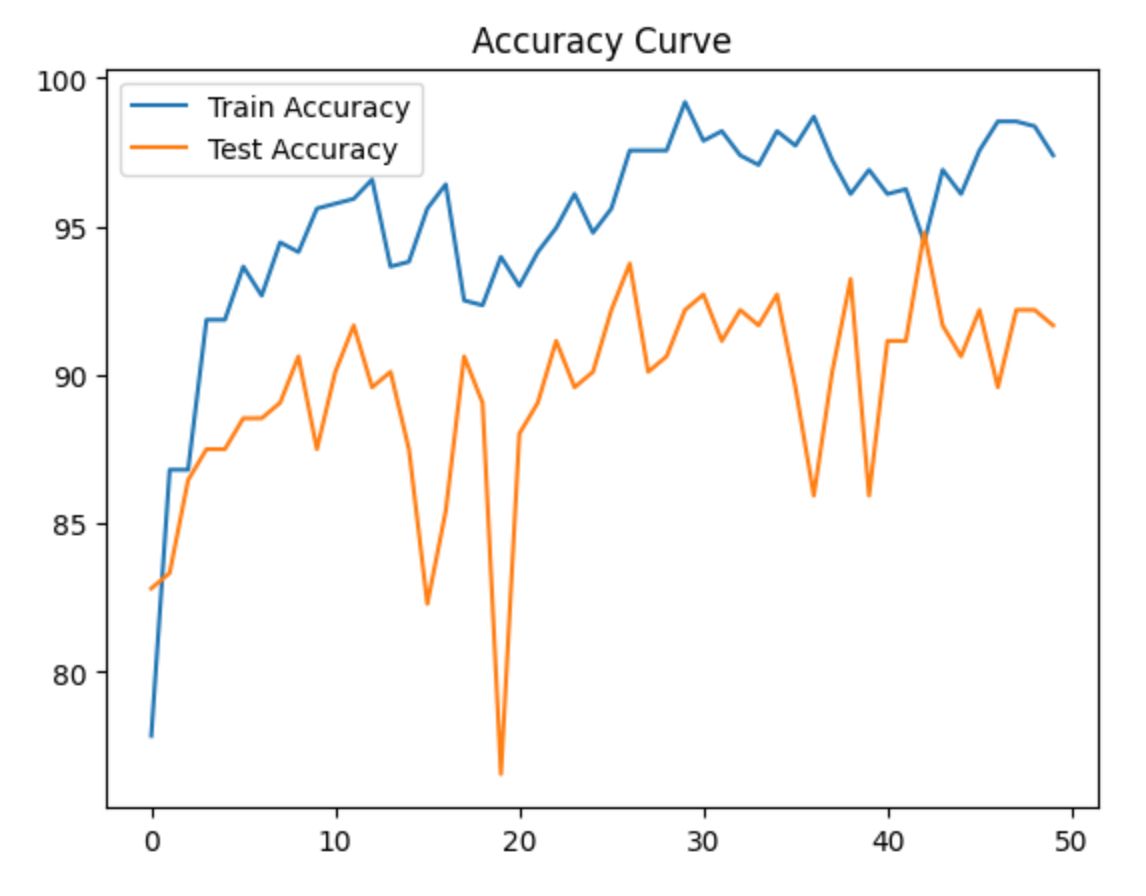}
    \caption{ The accuracy metrics over the range of 0 to 50 epochs over training and testing datasets.}
\end{figure}
From figure 3 it is visualizable that the training accuracy was excellent. A maximum training accuracy of 98\% to even 99\% was achieved by the model. Both training and testing accuracy were initially lower, but with consistent model improvisation, there is steady improvement seen along both parameters. Peak testing accuracy of 94.8\% was achieved, indicating superior model generalization despite mild fluctuations in the results during the testing process.

\begin{figure}[H]
    \centering
    \includegraphics[width=1.0\linewidth]{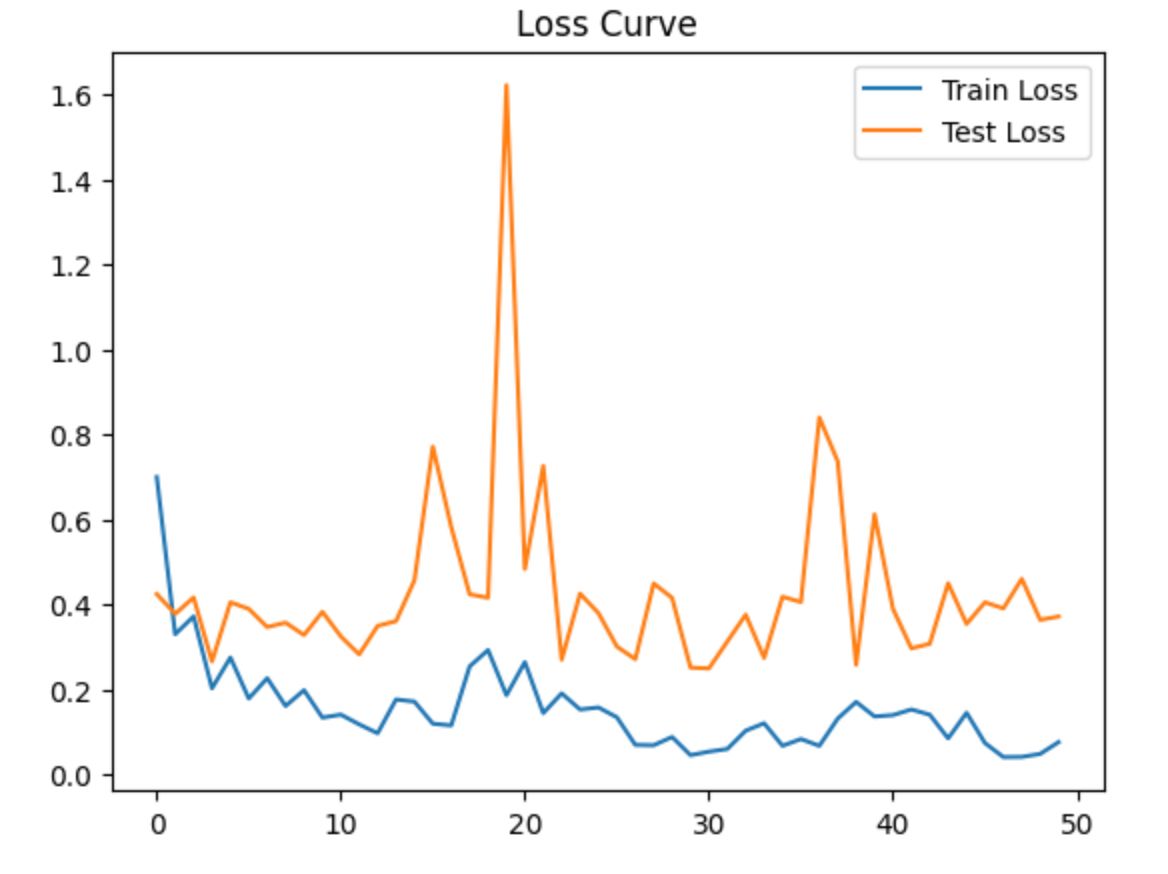}
    \caption{The loss values over the range of 0 to 50 epochs over training and testing datasets.}
\end{figure}

The hybrid model is considerably deep and robust with lots of hyperparameter tweaking, ultimately leading it to mild overfitting. Mostly imbalanced datasets are only available in this domain. The hybrid dataset that is made use of here has minor class imbalance. 

From the loss graphs, it is understandable that the training loss was high initially, but with time it reduced significantly, which is a strong positive sign. However, the testing loss remained higher than the training loss and was also fluctuating around epochs 20 and 35. This indicates mild overfitting that is hard to eliminate  despite heavy augmentation followed by dropout and regularization layers due to outliers in the data with minor imbalance among classes. 

\subsection{Ablation studies}
\begin{table}[htbp]
\centering
\caption{Results of Ablation Study}
\begin{tabular}{|l|c|}
\hline
\textbf{Model Variant} & \textbf{Accuracy (\%)} \\ \hline
\textbf{EfficientNet-B0 + ViT + Cross-Attention (Proposed)} & \textbf{94.79} \\ \hline
EfficientNet-B0 + ViT (Concatenation, No Attention) & 91.67 \\ \hline
EfficientNet-B0 + ViT + Self-Attention & 87.62 \\ \hline
Vision Transformer (ViT) Only & 86.72 \\ \hline
EfficientNet-B0 (CNN) Only & 85.98 \\ \hline
\end{tabular}
\end{table}
The core learnings from the detailed ablation study are stated in Table IV as a comprehensive snapshot. It is overtly evident that the final developed framework with both Vision Transformers and EfficientNet-B0, when fused with each other using the cross-attention module, seamlessly outclasses the other models with 94.79\% accuracy. Even while using a simple feature concatenation process, the model performs fairly well with a commendable 91.67\% accuracy. 

It's interesting to note that when self-attention is used instead of cross-attention, the model's performance dips quite a bit to 87.62\% accuracy, which implies fusing both the global and local features in a guided manner plays a pivotal role.  Both the Vision Transformer and EfficientNet-B0 models on their own performed satisfactorily with 86.72\% and 85.98\% accuracy.

\subsection{Data Visualization with Grad-CAM}\

\begin{figure}[H]
    \centering
    \includegraphics[width=0.70\linewidth]{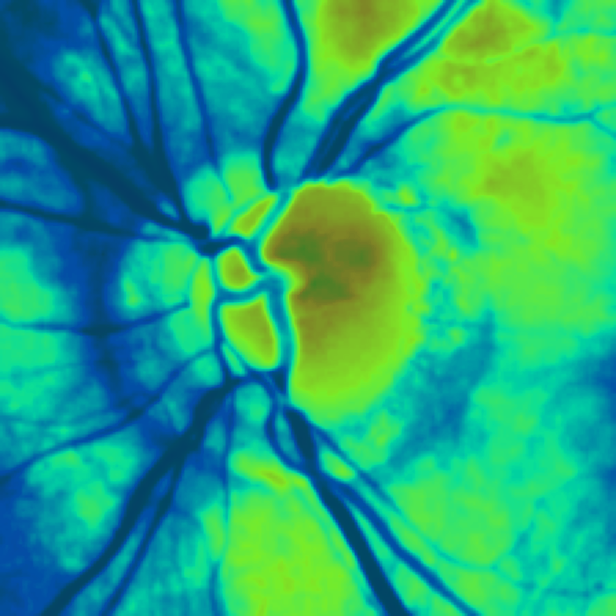}
    \caption{This image evidently highlights the optic disc and cup region, showing significant anomalies.}
    \label{fig:gradcam1}
\end{figure}

\begin{figure}[H]
    \centering
    \includegraphics[width=0.70\linewidth]{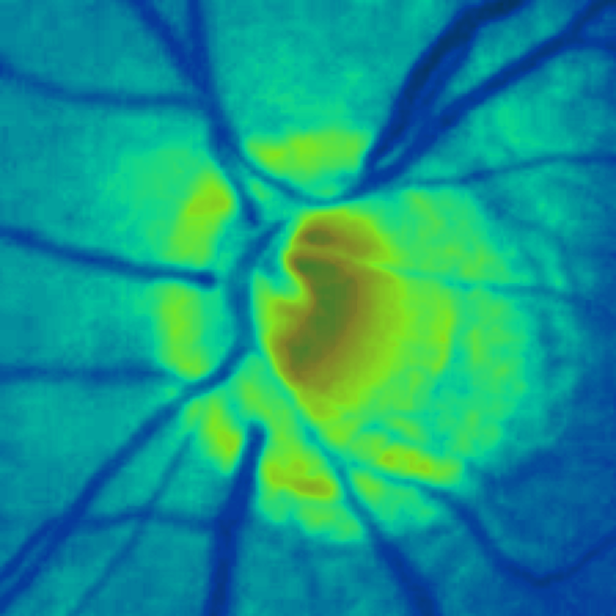}
    \caption{Optic disc and cup boundaries are extremely high, indicating advanced disease progression.}
    \label{fig:gradcam2}
\end{figure}

\begin{figure}[H]
    \centering
    \includegraphics[width=0.70\linewidth]{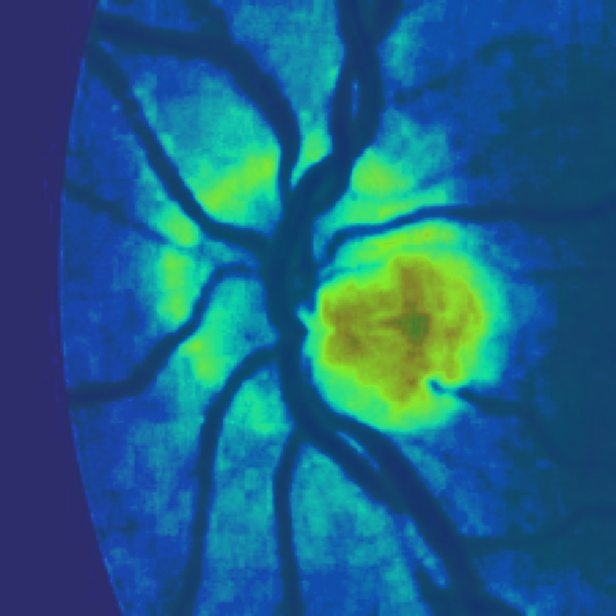}
    \caption{Image with no significant optic disc enlargement with thick and healthy nerve fibers.}
    \label{fig:gradcam3}
\end{figure}

\begin{figure}[H]
    \centering
    \includegraphics[width=0.70\linewidth]{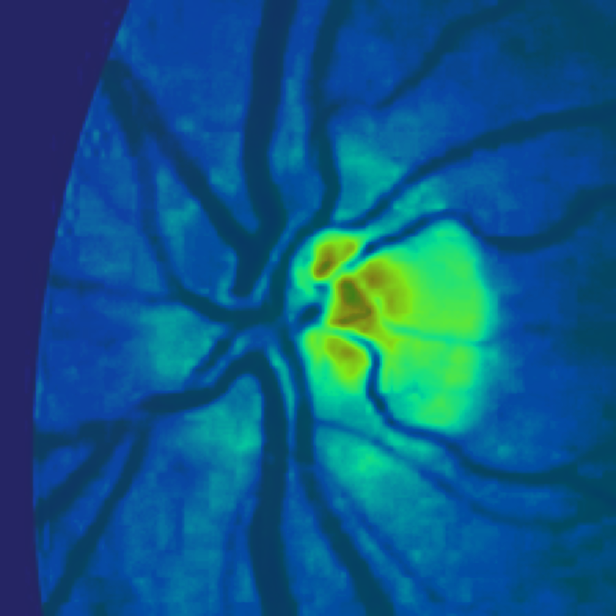}
    \caption{ A healthy eye with a normal disc-to-cup ratio, no cupping present}
    \label{fig:gradcam4}
\end{figure}

[18] Grad-CAM, which is Gradient-Weighted Class Activation Mapping put into existence by researchers from Georgia Tech, is a technique to explore the model's transparency and interpretability and a crucial part of explainable artificial intelligence (XAI). The figures ~\ref{fig:gradcam1},~\ref{fig:gradcam2}, ~\ref{fig:gradcam3} and ~\ref{fig:gradcam4} unequivocally demonstrate the diverse features of the fundus image via a heatmap representation that has played a crucial role in identifying the canonical traits of the disease.

Mainly in figure ~\ref{fig:gradcam1} and ~\ref{fig:gradcam2} it is seen that the optic disc along with the optic cup has enlarged in terms of dimensions. The neuroretinal rim has undergone excessive thinning to the point where nerve fibers are barely visible. Regarding the figures ~\ref{fig:gradcam3} and ~\ref{fig:gradcam4} the heat map seems to be less highlighted. It is due to the healthy optic disc and cup magnitudes. The ratio between the disc and cup is within acceptable limits for normal eyes. The nerve fibers, mainly in the rim area, are also blatantly seen, which is a signature marker of good eyes.

Due to the use of such a superlative utility to portray how the model learns and perceives various elements of the image, this can discernibly improve investor sentiment by making them readily invest in and fund such research. This also facilitates trust among doctors and reassures confidence in the patients. This will enable elevated real-time clinical usage, especially in developing countries internationally. But most of the AI research is typically black-box driven, meaning it is incomprehensible to a novice user.

\vspace{-15mm} 

\section*{Conclusion and Future Work}

It is highly anticipated that this research project will be deployed in an open-source, effortlessly accessible cloud platform like AWS and Google Cloud Platform, where real-time data can be tested using this model. An option for real-time training and improvisation of the model is also set to be made.

\end{document}